\documentclass{ieeetj}
\usepackage{cite}
\usepackage{amsmath,amssymb,amsfonts}
\usepackage{algorithm}
\usepackage{algpseudocode}  
\usepackage{graphicx,color,array}
\usepackage{textcomp}
\usepackage{xcolor}
\usepackage{hyperref}
\hypersetup{hidelinks=true}
\usepackage{booktabs}  
\usepackage{arydshln}
\usepackage{multirow} 
\def\BibTeX{{\rm B\kern-.05em{\sc i\kern-.025em b}\kern-.08em
    T\kern-.1667em\lower.7ex\hbox{E}\kern-.125emX}}
\AtBeginDocument{\definecolor{tmlcncolor}{cmyk}{0.93,0.59,0.15,0.02}\definecolor{NavyBlue}{RGB}{0,86,125}}

\def\OJlogo{\vspace{-4pt}$<$Society logo(s) and publication title will appear here.$>$}
\def\seclogo{\vspace{10pt}$<$Society logo(s) and publication title will appear here.$>$}

\def\authorrefmark#1{\ensuremath{^{\textbf{#1}}}}

\begin{document}
\receiveddate{XX Month, XXXX}
\reviseddate{XX Month, XXXX}
\accepteddate{XX Month, XXXX}
\publisheddate{XX Month, XXXX}
\currentdate{XX Month, XXXX}
\doiinfo{XXXX.2022.1234567}

\markboth{}{Author {et al.}}

\title{Privacy-Preserving Fair Synthetic Tabular Data}

\author{Fatima J. Sarmin\authorrefmark{1}, Atiquer R. Rahman\authorrefmark{1},\\ Christopher J. Henry \authorrefmark{1} (Senior Member, IEEE),\\ and Noman Mohammed \authorrefmark{1}}
\affil{Department of Computer Science, University of Manitoba, Manitoba, Canada}
\corresp{Corresponding author: Fatima J. Sarmin (email: sarminf@myumanitoba.ca).}
\authornote{ARS was a Gordon P. Osler scholar and was partially supported by UMGF fellowship. NM was supported by the NSERC Discovery Grants (RGPIN-04127-2022) and the NSERC Alliance Grants (ALLRP 592951 - 24).}

\begin{abstract}
Sharing of tabular data containing valuable but private information is limited due to legal and ethical issues. Synthetic data could be an alternative solution to this sharing problem, as it is artificially generated by machine learning algorithms and tries to capture the underlying data distribution. However, machine learning models are not free from memorization and may introduce biases, as they rely on training data. Producing synthetic data that preserves privacy and fairness while maintaining utility close to the real data is a challenging task. This research simultaneously addresses both the privacy and fairness aspects of synthetic data, an area not explored by other studies. In this work, we present PF-WGAN, a privacy-preserving, fair synthetic tabular data generator based on the WGAN-GP model. We have modified the original WGAN-GP by adding privacy and fairness constraints forcing it to produce privacy-preserving fair data. This approach will enable the publication of datasets that protect individual's privacy and remain unbiased toward any particular group. We compared the results with three state-of-the-art synthetic data generator models in terms of utility, privacy, and fairness across four different datasets. We found that the proposed model exhibits a more balanced trade-off among utility, privacy, and fairness.
\end{abstract}

\begin{IEEEkeywords}
Data privacy, data fairness, generative adversarial networks, synthetic data generation.
\end{IEEEkeywords}


\maketitle

\section{INTRODUCTION}
\IEEEPARstart{T}{o} keep pace with modern data-driven developments and artificial intelligence (AI) advancements, vast amounts of data are essential. The main objectives of sharing data are to obtain statistical information and to train AI models and software.  Tabular data is one of the most frequently utilized forms \cite{borisov2022deep}, as it is prevalent across various real-world domains. Examples of domains that utilize tabular data include healthcare with electronic health records, planning and development in governance using census data, web logs for cybersecurity, transaction logs for credit scoring and financial planning, and participant data for various scientific research \cite{fatima2017survey, buczak2015survey, dastile2020statistical, shwartz2022tabular}. Most often, real data is expensive, not always available, and must be handled with care, especially for sensitive information such as medical records or credit data, which contain personal information protected by data regulatory laws. To mitigate the legal and ethical risks of sharing data, de-identification of sensitive information is commonly used. However, previous research has demonstrated that when linked with other identifiable datasets, de-identification cannot fully prevent re-identification risks \cite{sweeney1997weaving, el2011re, malin2004not, erlich2014routes, sarkar2024identification}. Additionally, real data and its subsequent de-identification may contain biases, which could lead to unfair decision-making if used to train AI models \cite{angwin2019machine, o2017weapons,dastin2022amazon,tashea2017courts, lu2020gender,de2021stereotype,kadambi2021achieving}. This could lead to discrimination against certain races or individuals and undermine people's faith in machine learning and AI. 

In this context, researchers and practitioners view synthetic data as a promising approach for open data sharing as it is artificially generated data \cite{arnold2020really, bellovin2019privacy,xu2019modeling,choi2017generating,drechsler2010sampling,yale2019assessing,yale2019privacy, Sarmin2024-mi}. Generating synthetic data is not a new concept \cite{rubin1993statistical}; it involves mimicking the properties and structure of real data.  Early methods used statistical techniques such as Bayesian networks  \cite{young2009using} and Hidden Markov models \cite{ngoko2014synthetic}. More recent approaches utilizes machine learning models such as Generative Adversarial Networks (GANs)\cite{goodfellow2020generative} and variational auto-encoders (VAEs) \cite{kingma2013auto}. However, with recent advancements in deep learning, GAN-based models \cite{xu2018synthesizing, xu2019modeling, zhao2021ctab} have become popular for generating tabular datasets due to their ability to produce high-quality synthetic data that accurately reflects the complexities of real tabular data. GANs excel at generating such data by leveraging an adversarial training process that enhances the realism and fidelity of the outputs. This approach makes GANs effective at preserving the complex dependencies between features and handling diverse data types.
High-quality synthetic data can mitigate challenges associated with real data sharing in the following ways:
\begin{itemize}
    \item By accurately modeling the real data distribution, generative models can produce synthetic data on demand, thereby resolving issues related to data availability. Once trained, these models can generate as much data as needed.
    \item Because synthetic data is generated artificially and does not preserve a one-to-one relationship with real data, it offers enhanced protection for individual privacy.
    \item Furthermore, by modifying the models to ensure fair data generation and addressing any biases found in the real data, we can effectively address concerns related to fairness.
\end{itemize}

\subsection{Motivation}
It was initially believed that high-quality synthetic data \cite{alaa2022faithful} would be free from privacy concerns due to its synthetic nature. However, Shokri et al. \cite{shokri2017membership} found that machine learning models have a tendency to memorize training data. Deep neural network models are highly complex, and their adversarial training heavily depends on the training data. As a result, there is a risk of re-identification, as real samples could reappear in the generated data. Moreover, Hitaj et al. \cite{hitaj2017deep} introduced an active inference attack that can reconstruct the training data from the generated synthetic data, posing a significant risk to individual privacy. This has led to the incorporation of theoretically guaranteed differential privacy (DP) \cite{jordon2018pate,el2020practical,liu2022privacy,xie2018differentially} in the generation process. However, as DP adds extra noise to the samples to ensure privacy, the generated data loses utility, leading to privacy-utility trade-offs.

The issue of fairness in AI models complicates this scenario further and is another critical concern. Biases present in training datasets can cause machine learning models to produce unfair data. If this data is used in decision-making processes, the outcomes will be unfair. This phenomenon has been observed in various domains, including a criminal justice system and an employee selection process, where biased AI systems have made decisions that disproportionately affect certain groups, often exacerbating existing inequalities \cite{chouldechova2017fair, lambrecht2019algorithmic}. This is a relatively new research area and is currently gaining focus in the research community \cite{van2021decaf, rajabi2022tabfairgan}.

Thus, privacy and fairness are crucial considerations in the synthetic data generation process, alongside the usefulness of the data. However, these two important factors, privacy and fairness, have not yet been studied extensively together. This has led us to explore the following research questions in the context of synthetic data generation:

\begin{enumerate}
    \item Do existing privacy-preserving synthetic data generation models ensure fairness?
    \item Do existing synthetic data generation models claiming fairness ensure privacy?
    \item What will be the effect on utility if we incorporate both privacy and fairness in the synthetic data generation model?
\end{enumerate}

Our goal is to address these research questions and generate privacy-preserving, fair synthetic tabular data. This is highly beneficial for synthetic data research, as it will eliminate the privacy concerns raised by data regulatory laws. Additionally, when this data is used to train machine learning models, it will be free from any bias in the decision-making process toward individuals or groups. This will ensure that AI systems can be developed and deployed in a manner that protects individuals' rights and promotes social equity.

\subsection{Contributions}
Our contributions to addressing the privacy and fairness concerns of synthetic tabular data are as follows: 
\begin{enumerate}
    \item  We empirically tested the performance of three existing models—WGAN-GP (Wasserstein Generative Adversarial Network with Gradient Penalty) \cite{arjovsky2017wasserstein}, TabFairGAN \cite{rajabi2022tabfairgan}, and ADS-GAN \cite{yoon2020anonymization}—on utility, fairness, and privacy dimensions using four different datasets. Earlier studies did not evaluate performance along these three dimensions simultaneously. WGAN \cite{arjovsky2017wasserstein} generates synthetic data without explicit measures to ensure privacy and fairness. ADS-GAN \cite{yoon2020anonymization} provides explicit privacy guarantees, while TabFairGAN \cite{rajabi2022tabfairgan} focuses explicitly on fairness.
    \item We propose PF-WGAN, a privacy-preserving and fair synthetic tabular data generator. For this, we modified the WGAN-GP \cite{arjovsky2017wasserstein}. We incorporated \textit{identifiability} \cite{yoon2020anonymization} for privacy and \textit{demographic parity} \cite{zafar2017fairness} for fairness as components to the loss function alongside the generator's existing loss function to ensure privacy and fairness in the generated data in the generator of the WGAN-GP \cite{arjovsky2017wasserstein} architecture during model training. To the best of our knowledge, this approach has not previously been introduced for incorporating both privacy and fairness in synthetic tabular data generation research. While some models use multiple generators and discriminators to produce fair data, we utilized a single generator and discriminator to generate synthetic data, simplifying the architecture without compromising performance.
    \item We compared the utility, privacy, and fairness of the data generated by our model against three other models using four different datasets (more details in Section \ref{secResult}). Our model demonstrated a more balanced trade-off among utility, privacy, and fairness. For instance, it provided better privacy than WGAN and TabFairGAN, although it was less protective than the privacy-focused ADS-GAN. Conversely, our model’s accuracy and F1-score were significantly better than those of ADS-GAN. In terms of fairness, measured through \textit{demographic parity}, our model outperformed the other generators on three of the four datasets. 
    
\end{enumerate}

The rest of the paper is organized as follows. Section \ref{secRelatedwork} describes the works related to our study in terms of privacy, and fairness in synthetic data generation. We explain the terms and notation used in this study in Section \ref{secPrelimin}. The PF-WGAN framework and its theoretical properties are introduced in Section \ref{secFram}. Section \ref{secImplem} details the implementation, including descriptions of the datasets used in the experiments, data preprocessing steps, and model training procedures. Section \ref{secResult} presents the experimental results and evaluates the models. Finally, Section \ref{secConclusion} concludes the paper.

\begin{table*}[h!]
\centering
\caption{Summary of related work on GAN- based synthetic tabular data generation models. We are interested in: (1) the number of generators; (2) the number of discriminators; and whether the model has (3) explicit privacy provision; (4) explicit fairness provision.}

\begin{tabular}{ m{2.1cm}  m{1.0cm}  m{1.0cm}  m{0.6cm}  m{0.6cm}  m{7.0cm} m{0.1cm}}
\hline
\textbf{Model} & \textbf{1} & \textbf{2} & \textbf{3} & \textbf{4} & \textbf{Objective} \\
\hline
\hline

DPGAN \cite{xie2018differentially} & Single & Single & \checkmark & \text{\textbf{X}} & Pivacy-preserving synthetic data with \textit{Differential Privacy}.  \\ 
[6pt]
PATE-GAN \cite{jordon2018pate} & Single & Multiple & \checkmark & \text{\textbf{X}}  & Privacy-preserving synthetic data with \textit{Differential Privacy}.  \\ 
[6pt]
ADS-GAN \cite{yoon2020anonymization} & Single & Single & \checkmark & \text{\textbf{X}} & Privacy-preserving synthetic data with \textit{Distance based metric}.  \\ 
[6pt]
FairGAN \cite{xu2018fairgan} & Single & Dual & \text{\textbf{X}} & \checkmark & Realistic \& fair synthetic data using \textit{Demographic Parity}. \\ 
[6pt]
CFGAN \cite{xu2019achieving} & Dual & Dual & \text{\textbf{X}} & \checkmark & Realistic \& fair synthetic data using \textit{Causal Intervention-Based Fairness}. \\ 
[6pt]
DECAF \cite{van2021decaf} & Multiple & Single & \text{\textbf{X}} & \checkmark & Realistic \& fair synthetic data using \textit{Causal Structure-Based Fairness}.  \\ 
[6pt]
TabFairGAN \cite{rajabi2022tabfairgan} & Single & Dual  & \text{\textbf{X}} & \checkmark & Realistic \& fair synthetic data with \textit{Demographic parity} in two step training. \\ 
\hline
\hline

\textbf{PF-WGAN (ours)} & Single & Single & \checkmark &\checkmark  & Realistic, privacy-preserving \& fair synthetic data.\\ 
\hline
\end{tabular}
\label{tab:all_models}
\end{table*}

\section{Related Works}\label{secRelatedwork}
The main purpose of this work is to find a technique to impose a balance between utility, privacy, and fairness in synthetic data generation. Initially, research in synthetic data generation focused primarily on creating realistic data without specific considerations for privacy and fairness. Some of the popular GAN-based models for tabular data generation include TGAN \cite{xu2018synthesizing}, CTGAN \cite{xu2019modeling}, CopulaGAN \cite{SDV}, and CTAB-GAN \cite{zhao2021ctab}. While these models excel at generating realistic data with intricate architectures and training processes, they fall short in preserving individual privacy and ensuring fairness. Later, researchers began addressing privacy concerns to protect sensitive information and comply with data publishing regulations. More recently, some researchers have also aimed to address unfairness in generated data. More recently, some researches are also investigating and addressing the fairness concerns in generated data. In subsection \ref{privmodel}, we discuss the models focusing on the privacy aspect and in subsection \ref{fairmodel}, we discuss the models focusing on the fairness aspect. However, most existing research focuses either on privacy-preserving realistic synthetic data generation or on fair and realistic synthetic data generation, but not both. None of these studies address all three aspects—utility, privacy, and fairness—together in synthetic data generation.

In our work, we aim to address these three issues simultaneously, enabling the sharing of synthetic tabular data that is both privacy-preserving and free from unfairness toward certain groups. Table \ref{tab:all_models} provides an overview of related works that focus exclusively on GAN-based synthetic data generation in terms of utility, privacy, and fairness. We categorize the models in the table according to our primary areas of interest: (1) number of generators; (2) number of discriminators; (3) privacy provision; and (4) fairness provision. We are particularly interested in the architecture of the models, as the model training time and complexity increase with the number of generators and discriminators used.

\subsection{Privacy-preserving models} \label{privmodel}
To use synthetic data as an alternative to real data, it must not only be realistic but also mitigate the legal and ethical risks associated with sharing sensitive information and protect against re-identification through linkage with other identifiable datasets. Privacy in synthetic data can be addressed using theoretical privacy guarantees, such as differential privacy \cite{dwork2014algorithmic, tao2021benchmarking,mckenna2108winning} or distance correlation-based methods \cite{yoon2020anonymization}. PATE-GAN \cite{jordon2018pate} and DPGAN \cite{xie2018differentially} are two popular differential privacy-based GAN methods that offer formal privacy guarantees. DPGAN \cite{xie2018differentially} adapts the GAN model to achieve differential privacy by adding noise to the discriminator's gradients and applying the Post-Processing Theorem, ensuring the generated data is differentially private. PATE-GAN \cite{jordon2018pate} achieves differential privacy by modifying the training procedure of the discriminator to be differentially private, using a modified version of the Private Aggregation of Teacher Ensembles (PATE), which involves multiple teacher models as discriminators, thus increasing model complexity. Introducing differential privacy in GANs typically decreases utility in both DPGAN \cite{xie2018differentially} and PATE-GAN \cite{jordon2018pate}. Another privacy-preserving model, ADS-GAN \cite{yoon2020anonymization}, generates synthetic data conditioned on the original data, with conditioning variables optimized using a conditional GAN model. Unlike differential privacy models, ADS-GAN \cite{yoon2020anonymization} employs a distance-based privacy metric, known as \textit{identifiability} to maintain better utility while still providing privacy. They have demonstrated that they provide privacy using distance-based methods, and their approach is better in terms of utility compared to PATEGAN \cite{jordon2018pate} and DPGAN \cite{xie2018differentially}, as adding extra noise to achieve differential privacy reduces utility. However, these models primarily focus on privacy and utility, not on fairness.

\subsection{Fairness-based Generation} \label{fairmodel}
Fairness in synthetic data is a less explored area than utility and privacy. FairGAN \cite{xu2018fairgan} is one of the earlier GANs that produces fair synthetic tabular data. In FairGAN \cite{xu2018fairgan}, there is one generator and two discriminators: one discriminator aims to ensure realistic generation, and the other aims to ensure fairness using demographic parity. CFGAN \cite{xu2019achieving} is another fair data generator model designed to reflect the structures of causal and interventional graphs. It has two generators and two discriminators. DECAF \cite{van2021decaf} is also a structural causal GAN model that allows each variable to be reconstructed conditioned on its causal parents. They used $d$ generators (one for each variable). Each variable is sequentially generated by its corresponding generator, utilizing parental information provided by the governing Directed Acyclic Graph (DAG) during the training. They removed the edge between the sensitive attribute and the target output to produce fair data. However, removing edges is too drastic and may result in unrealistic data. CFGAN \cite{xu2019achieving} and DECAF \cite{van2021decaf} rely heavily on the accuracy of the causal graph, which makes them less scalable to larger datasets with complex interdependencies. Recently, another GAN model called TabFairGAN \cite{rajabi2022tabfairgan} also produces fair tabular data through two-phase training. In the first phase, they train their model for accuracy. They define a separate critic network that works in the second phase to check the fairness in the generated data. 
\begin{figure*}[t]
  \centering
  \graphicspath{{Figure/}}
  \includegraphics[ width= 1\linewidth]{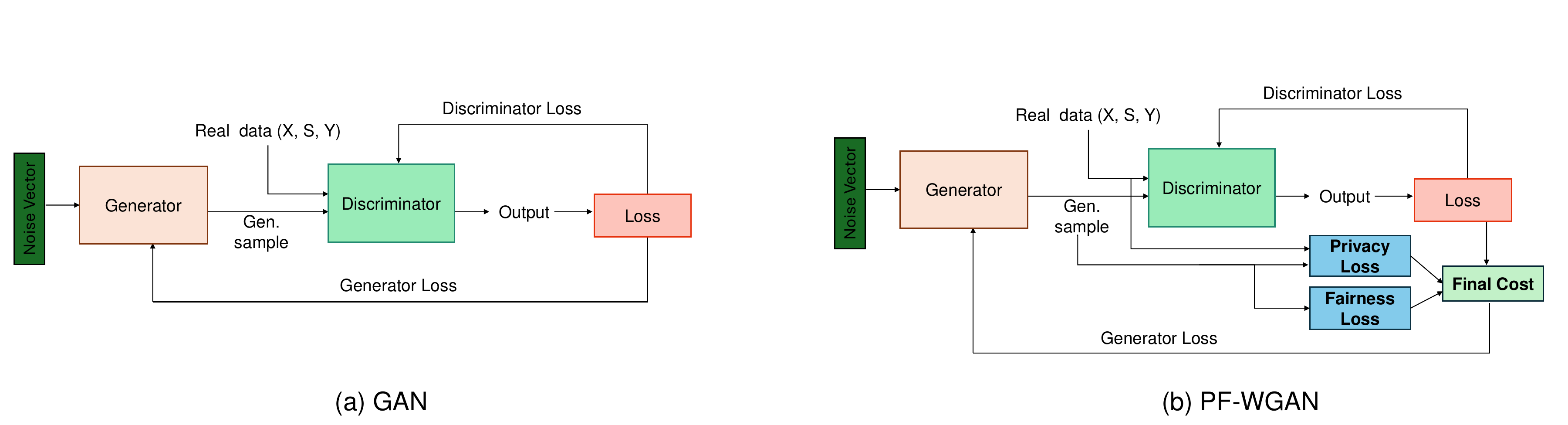}
  \caption{(a) Basic Generative Adversarial Networks (GANs) Architecture; (b) Proposed Network Architecture (Privacy preserving Fair WGAN: PF-GAN)}\label{fig:GAN-PFWGAN}
\end{figure*}
All these models either use two or more generators or discriminators or employ multiple training phases to generate fair synthetic data. The use of multiple generators or discriminators increases the model's complexity for larger datasets. Additionally, these models do not address privacy concerns. 

In our study, we address the gaps in previous research. To overcome these challenges, we use a simplified WGAN-GP model, which is more stable than original GANs, with only one generator and one discriminator. To ensure privacy and fairness, we incorporate \textit{identifiability} \cite{yoon2020anonymization} and \textit{demographic parity} \cite{zafar2017fairness} as additional loss functions alongside the generator’s original loss during training. Using \textit{identifiability} \cite{yoon2020anonymization} and \textit{demographic parity} \cite{zafar2017fairness} allows us to move away from reliance on differential privacy and causal graphs. In doing so, we aim to achieve a more practical and effective balance of privacy, fairness, and model simplicity in synthetic tabular data generation.

\section{Preliminaries}\label{secPrelimin}
Let, \( \mathcal{D}= \) \{\( \mathcal{X,S,Y} \)\} be a real tabular dataset with non sensitive variable, \( \mathcal{X} \), sensitive variable, \( \mathcal{S} \), and target outcome, \( \mathcal{Y} \). Here, sensitive variables are those values for which we want to ensure fairness in the target outcome.

Our goal is to generate a synthetic tabular dataset, \(\hat{\mathcal{D}} = \{\hat{\mathcal{X}}, \hat{\mathcal{S}}, \hat{\mathcal{Y}}\}\), where generated data will be privacy-protected and fair.

\subsection{Privacy}
We can say a synthetic dataset \(\hat{\mathcal{D}}\) is privacy-preserving if no real data sample appears in the synthetic dataset \(\hat{\mathcal{D}}\) and if the generated synthetic samples are different enough from the real data samples. To ensure privacy in datasets, differential privacy is widely used in computer science \cite{dwork2014algorithmic, mckenna2021winning}. However, incorporating differential privacy into a GAN-based model introduces extra noise, which can result in a loss of utility \cite{yoon2020anonymization, tao2021benchmarking}. In contrast, distance-based methods has been shown to preserve better utility while still providing privacy \cite{yoon2020anonymization}. In ADS-GAN \cite{yoon2020anonymization}, privacy is defined in terms of \textit{identifiability}. 

\textit{Identifiability score} $\mathcal{I}$ $(D, \hat{D})$ of a synthetic datasets $\hat{D}$ with respect to the original dataset $D$ is calculated as the ratio of synthetic records whose distance to the closest real point $(d)$ is less than the distance of the nearest real neighbor of that real point $(\hat{d})$. The formula is defined as follows: 

\begin{equation}
    \mathcal{I}(D, \hat{D}) = \frac{1}{N} \left[ \mathbb{I} \left( \hat{d}_i < d_i \right) \right]
\end{equation}
where $\mathbb{I}$ represents the identity function and
\[
d_i = \min_{x_j \in D / x_i} \left\| w \cdot (x_i - x_j) \right\|
\]
\[
\hat{d}_i = \min_{\hat{x}_j \in \hat{D}} \left\| w \cdot (x_i - \hat{x}_j) \right\|
\]

where $x$ is the real data sample and $\hat{x}$ is the generated synthetic sample. In this work, we used \textit{identifiability} as the measure of privacy. It is well-known and the \textit{identifiability score} is widely used in other research articles as well \cite{lautrup2025syntheval, liu2024exploring, lautrup2024systematic, shi2022generating, hashemi2023time, miletic2024assessing}.

\subsection{Fairness}
Fairness ensure that the probability of a favorable outcome (e.g., being hired, receiving a loan) is independent of a sensitive attributes (e.g., race, gender). There are different studies of fairness on dataset and classification in the literature \cite{hardt2016equality, kamiran2012data}. Among them, we chose \textit{Demographic Parity} \cite{zafar2017fairness}, one of the popular definitions for defining fairness in a labeled dataset, for our purpose.

For a given labeled dataset \( \mathcal{D}\), demographic parity or statistical parity is define by:
\begin{equation}
P(Y = 1 \mid S = 1) = P(Y = 1 \mid S = 0).
\end{equation}
This is a fairness criterion in machine learning and decision-making processes. Formally, a decision-making algorithm satisfies \textit{demographic parity} if the likelihood of a positive decision is the same across different groups defined by the sensitive attribute. For example, in adult dataset, if the sensitive attribute is gender (\( \mathcal{S}={Male, Female} \)) and the target outcome (\( \mathcal{Y}={(\leq 50\text{K}, >50\text{K})} \)) is income then \textit{demographic parity} ensure that the income should not depend on the person's gender. In other words, \textit{demographic parity} is achieved if the probability of obtaining a favorable outcome (e.g., earning $>50\text{K}$) is the same for both males and females in the dataset. In the context of synthetic data, a generated synthetic dataset \(\hat{\mathcal{D}}\) is considered fair if it follows the \textit{demographic parity} equation, meaning the probability of a favorable outcome (e.g., earning $>50\text{K}$) is equal for both males and females. 

\section{PF-WGAN Framework}\label{secFram}
The basic principle of GAN models is shown in Fig.\ref{fig:GAN-PFWGAN} (a). A basic GAN model consists of two neural networks. The first one is called the \textit{Generator}, whose job is to produce fake or synthetic data from a noise vector. The second neural network is called the \textit{Discriminator}, whose job is to determine if the input is real (from the training set) or fake (produced by the \textit{Generator}). During training, a cost is calculated from the output of the \textit{Discriminator} and fed back to both networks to improve their performance. The \textit{Generator} tries to fool the \textit{Discriminator} by producing more realistic data.

Traditional GANs have some limitations, such as mode collapse, training instability, and the vanishing gradient problem. WGAN-GP \cite{arjovsky2017wasserstein}, an improved version of the original GAN, addresses these problems and provides solutions by using the \textit{‘Wasserstein distance’} (also known as Earth Mover's distance \cite{rubner2000earth}) as a measure of similarity between the true data distribution and the generated distribution. The wasserstein distance provides a more stable and meaningful measure of the discrepancy between distributions compared to traditional GAN objectives like the \textit{Jensen-Shannon divergence} \cite{menendez1997jensen}. This prevents the basic problems described above. Convergence properties and Lipschitz continuity also give theoretical guarantees to the WGAN-GP \cite{arjovsky2017wasserstein} network for more stable training.

The discriminator's primary role is to distinguish between real and generated data samples, acting as a binary classifier in a traditional GAN, and the discriminator's output does not directly measure the similarity between the distributions of real and generated data. On the other hand, the discriminator in WGAN-GP \cite{arjovsky2017wasserstein} evaluates the quality of the generated samples in a continuous manner instead of outputting a binary classification, acting as a critic. That’s why the discriminator in WGAN-GP \cite{arjovsky2017wasserstein} is called the \textit{‘Critic’}. We have taken WGAN-GP \cite{arjovsky2017wasserstein} with a gradient penalty as the base model.

The Fig.\ref{fig:GAN-PFWGAN} (b) illustrates the network architecture of the proposed model. Here we have incorporated privacy and fairness as loss functions with the WGAN-GP \cite{arjovsky2017wasserstein} loss function. The equation for the final loss function is as follows:
\begin{equation}
L_{final}=l_{WGAN-GP} + l_ {privacy} +l_ {fairness}.  
\end{equation} 
where $l_{WGAN-GP}$ is the loss from WGAN-GP \cite{arjovsky2017wasserstein}, $l_ {privacy}$ is the privacy loss, and $l_ {fairness}$ is the fairness loss. As we have modified the WGAN-GP \cite{arjovsky2017wasserstein} loss for the generator during training and added the privacy and fairness factors as new losses for the generator, we call it ‘Privacy-Preserving Fair WGAN (PF-WGAN).’

\begin{algorithm}[htbp]
\caption{PF-WGAN: Privacy and Fairness-Enhanced WGAN;
parameters: $\lambda$ = 10, $\lambda_{p}$ = 0.2, $\lambda_{f}$ = 1.0, $m$ = 256, $n_{\text{critic}}$ = 4, and Adam optimizer with $\alpha_{g}$ = 0.0001, $\alpha_{c}$ = 0.0002, $\beta_{1}$ = 0.5, $\beta_{2}$ = 0.999 }
\label{alg:modified_wgan}
\begin{algorithmic}[1]
\Require $\alpha_{g}$: the generator learning rate, $\alpha_{c}$: the critic learning rate, ($\beta_{1}$, $\beta_{2}$): decay rates, $\lambda$ : the gradient penalty coefficient, $\lambda_{p}$ : privacy loss weight, $\lambda_{f}$ : fairness loss weight, $m$: batch size, $n_{\text{critic}}$: number of critic iterations per generator iteration, $E$: total number of epochs, ($PF_{start}, PF_{end}$) : beginning and ending of adding privacy and fairness loss between the total number of epochs. 

\For{$i = 1, \ldots, E$}
    \For{$t = 1, \ldots, n_{\text{crit}}$}
        \State Sample a batch of size, $m$:
        \State {\scriptsize $D(x, y, s) \sim P_r$, $z \sim P(z)$, and $e \sim U[0, 1]$}
        \State {\scriptsize $\hat{D} = (\hat{x}, \hat{s}, \hat{y}) \leftarrow G_\theta(z)$}
        \State {\scriptsize $\bar{D} \leftarrow eD + (1 - e)\hat{D}$}
        \State Update the critic:
        \State \quad{\scriptsize$\nabla_w \left(\frac{1}{m} \sum_{i=1}^m C_w(\hat{D}) - C_w(D) + \lambda(||\nabla_{\bar{D}} C_w(\bar{D})||_2 - 1)^2\right)$}
    \EndFor
    \State Sample a batch of size $m$: {\scriptsize $\hat{D} = (\hat{x}, \hat{s}, \hat{y}) \sim P(G_\theta(z))$}
    \State Update the generator:
    \State \quad {\scriptsize $\nabla_\theta \left(\frac{1}{m} \sum_{i=1}^m -C_w(\hat{D})\right)$}

    \If{$PF_{start} < i < PF_{end}$}
        \State Calculate privacy loss:
        \State \quad {\scriptsize $l_{\text{privacy}} = \lambda_p ||w \cdot (D_k - \hat{D}_k)||$}
        \State Calculate fairness loss: 
        \State \quad {\scriptsize $l_{\text{fairness}} = \lambda_f \left( \frac{|D_{s=0, y=1}|}{|D_{s=0}|} - \frac{|D_{s=1, y=1}|}{|D_{s=1}|} \right)$}
        \State Update the generator with privacy and fairness loss:
        \State \quad {\scriptsize $\nabla_\theta \left(\frac{1}{m} \sum_{i=1}^m -C_w(\hat{D}) + l_{\text{privacy}} + l_{\text{fairness}} \right)$}
    \EndIf
\EndFor
\end{algorithmic}
\end{algorithm}

Algorithm \ref{alg:modified_wgan} summarizes the PF-WGAN's training process. We keep the critic the same as the original WGAN-GP critic. We modified the loss during generator training.

Our main modification is shown in lines (14-19). First, we calculate the original WGAN-GP accuracy loss to update the generator loss. Then, we calculate the privacy loss and fairness loss. We add these losses to the generator's original loss and update the generator loss so that it can produce privacy-preserving fair data. We adopted the \textit{identifiability} concept of ADS-GAN \cite{yoon2020anonymization} for the privacy loss function, mentioned in the equation \ref{eq:privloss}. 
\begin{equation}
l_{\text{privacy}} = \lambda_p ||w \cdot (D_k - \hat{D}_k)||.
\label{eq:privloss}
\end{equation}
The privacy loss $(l_{\text{privacy}})$ penalizes generated samples that are too similar to real samples. We experimented with two different variations of the privacy loss stated in equation  \ref{eq:privloss}.
\begin{enumerate}
    \item For the first approach $(l_{\text{privacy-1}},\text{henceforth L1})$, the mean squared distance between each generated sample $(\hat{D}_{k})$ and its corresponding real sample $(D_{k})$ is calculated.
    \item For the second approach $(l_{\text{privacy-NN}}, \text{henceforth L2})$, it is computed using the nearest-neighbor distance between each generated sample $(\hat{D}_{k})$ to its closest real sample $(D_{k})$ (nearest neighbor). This method is more comprehensive than the previous one and also more computationally expensive. Due to its computational cost, ADS-GAN \cite{yoon2020anonymization} approximates this loss using the $(l_{\text{privacy-1}})$ loss. By minimizing the privacy loss, the generator attempts to ensure that the generated samples do not exactly replicate real data and maintain some distance, thereby improving privacy preservation.
\end{enumerate}

We use \textit{demographic parity} to define fairness loss functions.
When a set of synthetic data is generated by the generator during training, we calculate the demographic parity for the privileged and unprivileged groups. The difference between these two groups is counted as the fairness loss, and we add this fairness loss function to the generator's loss function to encourage the generator to produce fair data in the future. We calculate the demographic parity as follows:

\begin{equation}
l_{\text{fairness}} = \lambda_f \left( \frac{|\hat{D}_{s=0, y=1}|}{|\hat{D}_{s=0}|} - \frac{|\hat{D}_{s=1, y=1}|}{|\hat{D}_{s=1}|} \right).
\label{eq:fairloss}
\end{equation}
Here $s$ represents the sensitive attribute, and $y$ represents the outcome of the corresponding synthetic data. The fairness loss measures the difference between the demographic parity rate of the privileged group and the underprivileged group. Adding this difference to the generator improves fair generation in future iterations.

\begin{table*}
\centering
\caption{Datasets Details}

\begin{tabular}{m{1.5cm} m{1.0 cm} m{1.0 cm} m{1.2 cm} m{1.2 cm} m{1.0 cm} m{1.4cm}}
\hline
\textbf{Datasets} & \textbf{Total records} & \textbf{Total Column} & \textbf{Numerical Column} & \textbf{Categorical Column} & \textbf{Sensitive Column} & \textbf{Output Column} \\
\hline
\hline
Adult & 48842 & 15 & 6 & 9 & Sex & Income \\
ProPublica & 16267 & 16 & 4 & 12 & Ethnicity & Recidivism \\
Bank & 45211 & 17 & 6 & 11 & Age & Subscription  \\
Law school & 19567 & 8 & 5 & 3 & Ethnicity & GPA \\
\hline
\end{tabular}
\label{tab:my_dataset}
\end{table*}

\section{Implementation Details}\label{secImplem}

\subsection{Dataset}
For this work, we have used four different datasets, which are given in Table \ref{tab:my_dataset}. We keep the column and record numbers the same as TabFairGAN \cite{rajabi2022tabfairgan} to compare our results with theirs. Each dataset is a combination of numerical and categorical values. We defined sensitive values and output columns for each dataset to measure demographic parity for fairness purposes.

The first dataset is the Adult dataset \cite{misc_adult_2}, which consists of over 48K records. Some examples of its attributes are employment, education, age, and gender. It is well-known for its bias towards predicting higher income $(>50\text{K})$ for males. That is why we selected Gender (Male, Female) as the sensitive attribute and income $(\leq 50\text{K}, >50\text{K})$ as the output column.

The second dataset is the ProPublica dataset \cite{Propublica} from the COMPASS risk assessment system. This provides data on offenders from Broward County, such as their ethnicity, marital status, and sex, as well as a score for each person indicating how likely they are to re-offend (Recidivism), which is the target attribute. The COMPASS risk assessment system has been found to be biased towards African-Americans, which is why we selected ‘Ethnicity’ as the sensitive attribute.

The third dataset is the Bank Marketing dataset \cite{moro2014data}, which contains data from a Portuguese banking institution’s direct marketing campaign. It includes people's age, profession, marital status, housing situation, etc. Here, the output column is the subscription to the term deposit (Subscription), and the sensitive attribute is age, as younger people are more likely to sign up for a term deposit than older people. We categorize those over age 25 as older.

The last dataset is the Law School dataset \cite{edERICED469370}, which includes records of law students with their GPA, race, and LSAT score. The target attribute is a binary variable showing their first-year average grade (GPA). Here, ‘Ethnicity’ is the sensitive attribute, as it has been observed that white students tend to have higher GPAs than black students.

\subsection{Data Preprocessing}
Properly pre-processed data leads to more stable training, better performance, and more accurate synthetic data generation. Generating tabular data using GAN networks is more challenging task due to the various data types in a single table. Preprocessing data for model training is crucial to handle the complexities of tabular data. Our datasets contain numerical and categorical data. It is important to preserved mutual dependency between any pair of attributes. Specifically, categorical values need to be converted into a numerical format to preserve the actual distinctions among categories without imposing a false order. Moreover, we are adding privacy and fairness loss functions to the generator loss function to generate privacy-preserving, fair synthetic data, so we have used some preprocessing steps to prepare our datasets. First, we used quantile transformation to transform numerical features into a uniform distribution. Then, we used one-hot encoding for the categorical features. For calculating the fairness loss, the sensitive column and the output column are needed. Therefore, we defined them within each dataset.

\subsection{Model training}
The network architecture of the PF-WGAN model is shown in Fig. \ref{fig:GAN-PFWGAN}(b). We have one Generator and one Critic. The Generator has one input layer, multiple hidden layers, and one output layer. We use different layers to process the numerical and categorical values. We use one linear and one batch normalization layer to process the numerical features. We use the ReLU activation function here. For handling categorical features, we use a linear layer for each categorical feature, transforming the input into one-hot encoded probabilities. We use ‘gumbel softmax’ as the activation function here. In the output layer, we combine numerical and categorical outputs into a single vector. The input and hidden layers of the Critic contain one linear layer and Leaky ReLU activation function, and one output layer. The layers in both the Generator and Critic allow the model to handle the diverse features and complex distribution of tabular data efficiently. 
The main improvement of the model for providing privacy and fairness is in the enhancement of the generator's loss function. As we know, the generator initially generates data from a uniform distribution. It is also well known that increasing privacy can decrease the utility of generated data. Therefore, adding privacy as an extra loss initially would result in less useful data. Thus, we first trained the generator to achieve better accuracy for a few epochs and then calculated the privacy and fairness loss of the generated data, adding these to the generator's loss to improve it in terms of privacy and fairness. We also used privacy and fairness loss weights to regulate the level of privacy and fairness in the synthetic data. For the Adult dataset, we divided the dataset into a 90:10 ratio, trained the model with 90\% of the data for 230 epochs, and used 10\% to evaluate the synthetic data. For the remaining three datasets, we divided each dataset into an 80:20 ratio, trained the model with 80\% of the data for 200 epochs, and used 20\% to evaluate the synthetic data.

During the implementation, we used the Adam optimizer with a generator learning rate $\alpha_{g}$ = 0.0001, a critic learning rate $\alpha_{c}$ = 0.0002, decay rates: $\beta_{1}$ = 0.5, $\beta_{2}$ =0.999, batch size $m$ = 256, the gradient penalty coefficient $\lambda$ =10, privacy loss weight $\lambda_{p}$ =0.2, fairness loss weight $\lambda_{f}$ =1.

\textbf{Challenges in model training:} Training a GAN network to produce tabular data is challenging due to the mixed data types involved. Without proper steps, the model can become unstable, and adding privacy and fairness as loss functions to the generator makes it even more complex. Initially, the extra loss functions caused loss explosion in the generator, resulting in \textit{‘NaN’} values. This was due to the fairness loss (calculated from \textit{demographic parity} in equation \ref{eq:fairloss}), which involved a \textit{‘divide by zero’} problem as it calculated the ratio between privileged and unprivileged groups. To address these issues, we implemented several approaches. We applied gradient clipping to the privacy and fairness losses, ensuring they stayed within minimum and maximum values. To avoid \textit{‘divide by zero’} errors, we added very small values to the losses. Additionally, we used \textit{‘batch normalization’} in the generator to stabilize training by normalizing the input to each layer, preventing the gradients from becoming too large. These approaches made the model more stable. When the privacy and fairness losses were added from the first epoch, the generator focused on producing more private and fair data, but less useful realistic outputs. To solve this, we trained the generator with only WGAN-GP's original loss for a few epochs before adding the privacy and fairness losses. This approach produced more realistic synthetic data that is also private and fair. We also weighted the privacy and fairness losses, allowing users to control the level of privacy and fairness desired (1 meaning highest privacy and fairness, and 0 meaning none).

\textbf{Evaluation:} After completing the model training, we generated the same amount of synthetic data as the real dataset for evaluation. We evaluated the synthetic data in terms of utility, fairness, and privacy. For calculating utility, we measure machine learning performance. To do this, we first trained two different decision tree classifiers with real and synthetic data to compare their performance. Specifically, we trained the model with synthetic data and tested it with real data for all synthetic data produced by different generation models. We compared machine learning accuracy, F1 score, and AUC-ROC for both real and synthetic data. We repeated each experiment 10 times to get the average score of the classifiers. For calculating the fairness of the generated data, we measured the \textit{demographic parity} in the generated data. We checked the ratio of favorable outcomes for both privileged and underprivileged groups for the sensitive column in the datasets.
For determining privacy, we measured the re-identification risk as the \textit{identifiability} \cite{qian2024synthcity}, which measures the distance between the real and synthetic data samples to find whether any real data has appeared in the synthetic data.

The environment and resources used to implement the code are listed in Table \ref{tab:my_EnHW}.

\begin{table}
\centering
\caption{Environment and Hardware}
\begin{tabular}{ l | l }
\hline

\textbf{Development Environment} & \textbf{Hardware} \\
\hline
Python: 3.7.16 & Processor: i7-8700\\
 PyTorch: 1.13.1+cu117 & RAM: 16GB\\
Numpy: 1.21.6 & GPU: Titan V GPU (12GB)\\
Pandas: 1.3.5 & \\
Scikit-learn: 1.0.2 & \\

\hline
\end{tabular}

\label{tab:my_EnHW}
\end{table}

\section{Results}\label{secResult}
To evaluate our model, PF-WGAN, we compare its results with real data, data generated by the base model WGAN \cite{arjovsky2017wasserstein}, one fair data generation model (TabFairGAN) \cite{rajabi2022tabfairgan}, and one privacy-preserving data generation model (ADS-GAN) \cite{yoon2020anonymization} in terms of utility, fairness, and privacy. We experimented with all these models across four different tabular datasets: Adult, ProPublica, Bank, and Law School. The experimental results are summarized in Table \ref{tab:comparison}. Each metric reported is the average of 10 experimental runs, providing a comprehensive evaluation of the models’ performance.

\begin{table*}[h!]
\centering
\caption{Comparison of models across various datasets. Lower values indicate better performance for fairness and privacy.}
\begin{tabular}{ | l | l | l  l  l | l| l |}
\hline
\textbf{Datasets} & \textbf{Models} & \textbf{Accuracy $\uparrow$} & \textbf{F1 Score $\uparrow$} & \textbf{AUC-ROC $\uparrow$} & \textbf{Fairness $\downarrow$} & \textbf{Privacy $\downarrow$} \\
\hline

\multirow{4}{*}{Adult} 
 & Real & 81.50\% ± 0.42\% & 61.73\% ± 0.72\%  & 75.00\% ± 0.42\% & 0.07 ± 0.00  & --\\
 & WGAN & 77.53\% ± 0.37\% & 50.25\% ± 0.56\% & 66.90\% ± 0.34\% & 0.10 ± 0.00  & 21.68\% ± 0.00\% \\
 & TabFairGAN & 76.70\% ± 1.06\% & 51.17\% ± 0.94\% & 67.80\% ± 0.59\% & 0.10 ± 0.01  & 25.12\% ± 2.27\% \\
 & ADS-GAN & 76.98\% ± 2.36\% & 35.54\% ± 11.98\% & 60.52\% ± 4.71\% & 0.96 ± 0.00  & 0.01\% ± 0.00\% \\
 & \textbf{Ours (L1)} & 71.54\% ± 0.68\% & 43.52\% ± 1.23\% & 62.81\% ± 0.83\% & 0.08 ± 0.00 & 17.82\% ± 0.00\% \\
 & \textbf{Ours (L2)} & 75.77\% ± 0.34\% & 47.33\% ± 0.76\% & 65.46\% ± 0.48\% & 0.11 ± 0.00 & 20.10\% ± 0.00\% \\
\hline

\multirow{4}{*}{ProPublica} 
 & Real & 90.48\% ± 0.24\%  & 91.51\% ± 0.27\% & 92.73\% ± 0.30\% & 0.26 ± 0.00  & -- \\
 & WGAN & 89.04\% ± 0.30\% & 90.22\% ± 0.29\% & 88.65\% ± 0.27\% & 0.26 ± 0.00 & 2.39\% ± 0.00\% \\
 & TabFairGAN & 87.65\% ± 1.43\% & 88.65\% ± 1.46\% & 87.51\% ± 1.36\% & 0.18 ± 0.04  & 2.61\% ± 0.07\% \\ 
 & ADS-GAN & 74.31\% ± 3.79\% & 71.32\% ± 6.00\% & 75.56\% ± 3.40\% & 0.20 ± 0.00 & 0.01\% ± 0.00\% \\
 & \textbf{Ours (L1)} & 89.13\% ± 0.55\% & 90.40\% ± 0.49\% & 88.65\% ± 0.55\% & 0.21 ± 0.00 & 2.53\% ± 0.00\% \\
 & \textbf{Ours (L2)} & 89.19\% ± 0.60\% & 90.45\% ± 0.53\% & 88.70\% ± 0.60\% & 0.23 ± 0.00 & 2.54\% ± 0.00\% \\
\hline

\multirow{4}{*}{Bank} 
 & Real & 87.69\% ± 0.34\% & 48.40\% ± 1.73\%  & 70.91\% ± 1.09\% & 0.026 ± 0.00  & -- \\
 & WGAN & 84.86\% ± 0.29\% & 33.41\% ± 0.98\% & 61.78\% ± 0.55\% & 0.02 ± 0.00  & 25.84\% ± 0.00\% \\
 & TabFairGAN & 85.08\% ± 0.62\% & 35.46\% ± 2.90\% & 63.29\% ± 1.57\% & 0.026 ± 0.02  & 30.42\% ± 0.01\% \\
 & ADS-GAN & 78.24\% ± 0.00\% & 29.91\% ± 0.04\% & 60.57\% ± 0.00\% & 0.12 ± 0.00  & 0.02\% ± 0.00\% \\
 & \textbf{Ours (L1)}  & 81.78\% ± 0.24\% & 33.12\% ± 1.03\% & 62.70\% ± 0.70\% & 0.02 ± 0.00  & 24.64\% ± 0.00\% \\
 & \textbf{Ours (L2)} & 82.61\% ± 0.39\% & 33.93\% ± 0.52\% & 62.94\% ± 0.43\% & 0.00 ± 0.00 & 25.33\% ± 0.00\% \\
\hline

\multirow{4}{*}{Law} 
 & Real & 85.57\% ± 0.45\% & 91.94\% ± 0.26\% & 61.95\% ± 0.82\% & 0.042 ± 0.00  & -- \\
 & WGAN & 82.29\% ± 0.47\% & 89.90\% ± 0.31\% & 61.56\% ± 1.23\% & 0.05 ± 0.00  & 17.26\% ± 0.00\% \\
 & TabFairGAN & 84.01\% ± 1.58\% & 91.08\% ± 0.95\% & 56.88\% ± 1.96\% & 0.005 ± 0.07  & 15.69\% ± 0.03\% \\
 & ADS-GAN & 78.11\% ± 2.73\% & 87.47\% ± 1.86\% & 49.80\% ± 1.64\% & 0.10 ± 0.00  & 0.07\% ± 0.00\% \\
 & \textbf{Ours (L1)}  & 87.08\% ± 0.44\% & 93.01\% ± 0.26\% & 53.26\% ± 0.72\% & 0.05 ± 0.00  & 14.96\% ± 0.00\% \\
 & \textbf{Ours (L2)} & 86.63\% ± 0.82\% & 92.74\% ± 0.48\% & 53.42\% ± 0.48\% & 0.03 ± 0.00 & 13.02\% ± 0.00\% \\
\hline

\end{tabular}
\label{tab:comparison}
\end{table*}

\begin{figure}[t]
  \centering
  \graphicspath{{Figure/}}
  \includegraphics[width=1\linewidth]{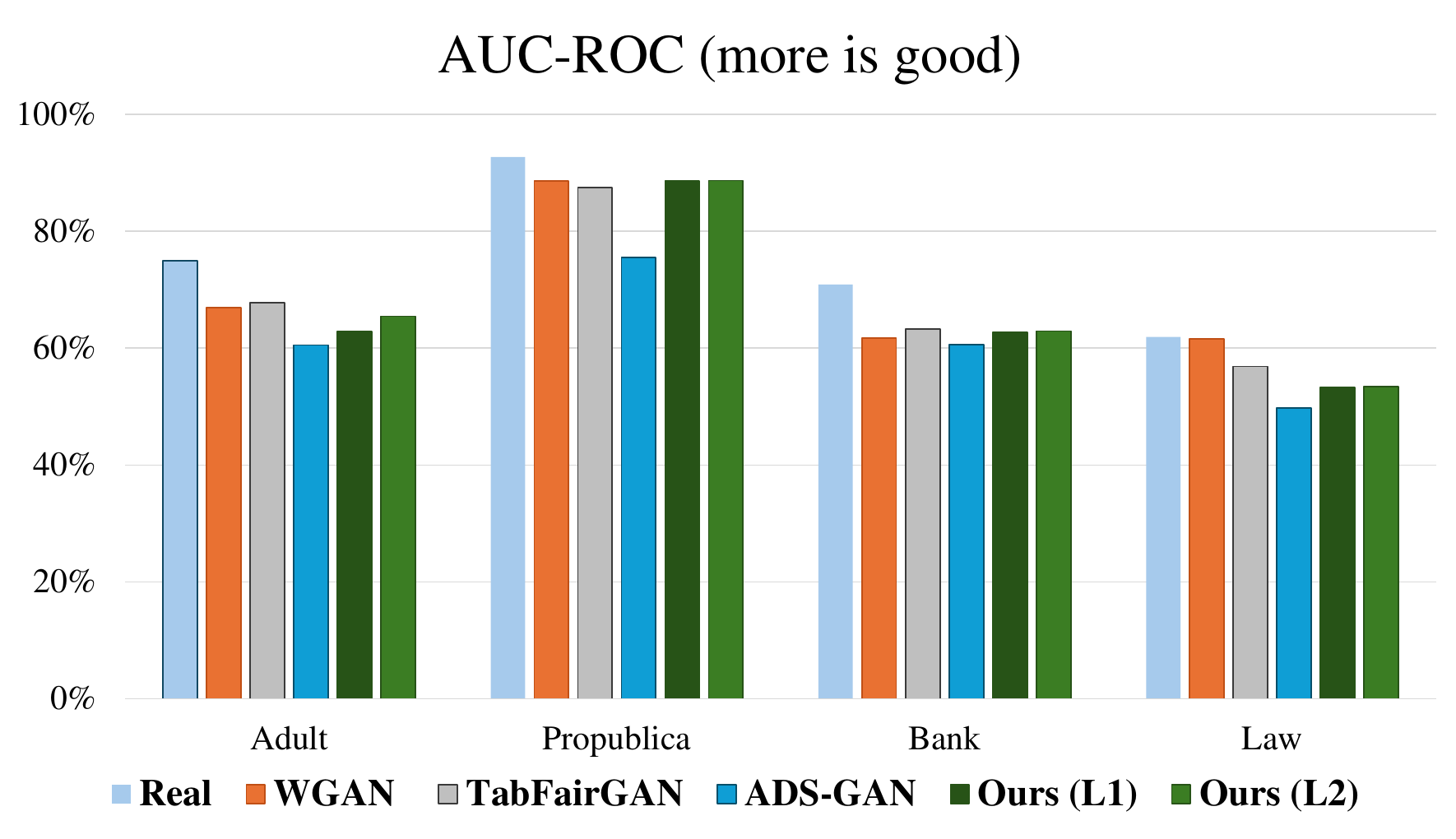}
  \caption{Result: Comparison among different models for utility (AUC-ROC score) using different datasets.}\label{fig:utility}
\end{figure}

We measure accuracy, F1 score, and AUC-ROC to evaluate the efficacy of our machine learning model. In terms of utility, our model, PF-WGAN, exhibits competitive performance with improvements in stability, as evidenced by lower standard deviations. For example, on the Adult dataset, PF-WGAN$(L2)$ achieves an accuracy score of 75.77\% ± 0.34\%, which is slightly lower in aggregate value than TabFairGAN (76.70\% ± 1.06\%) but shows a lower standard deviation. This indicates that our model is more stable than TabFairGAN. Similarly, on the Bank dataset, while PF-WGAN's utility in terms of accuracy, F1, and AUC-ROC is slightly lower than that of WGAN-GP and TabFairGAN, the standard deviation of our model is considerably lower, highlighting its more consistent performance. This trend is consistent across other datasets, such as the Law dataset, where our model PF-WGAN$(L1)$ achieves a higher F1 score of 93.01\% ± 0.26\% and a lower standard deviation compared to TabFairGAN, indicating improved robustness in its utility metrics. However, on the ProPublica dataset, our model achieves the best accuracy, F1 score, and AUC-ROC. Figure \ref{fig:utility} shows a comparison of these models' utility performance in terms of AUC-ROC for the all datasets. In three of the four datasets used in the experiments, the different versions of privacy loss used in our proposed model produced similar results, supporting ADS-GAN’s observation that the nearest-neighbor distance can be approximated by the corresponding paired distance.

\begin{figure}[t]
  \centering
  \graphicspath{{Figure/}}
  \includegraphics[width=1\linewidth]{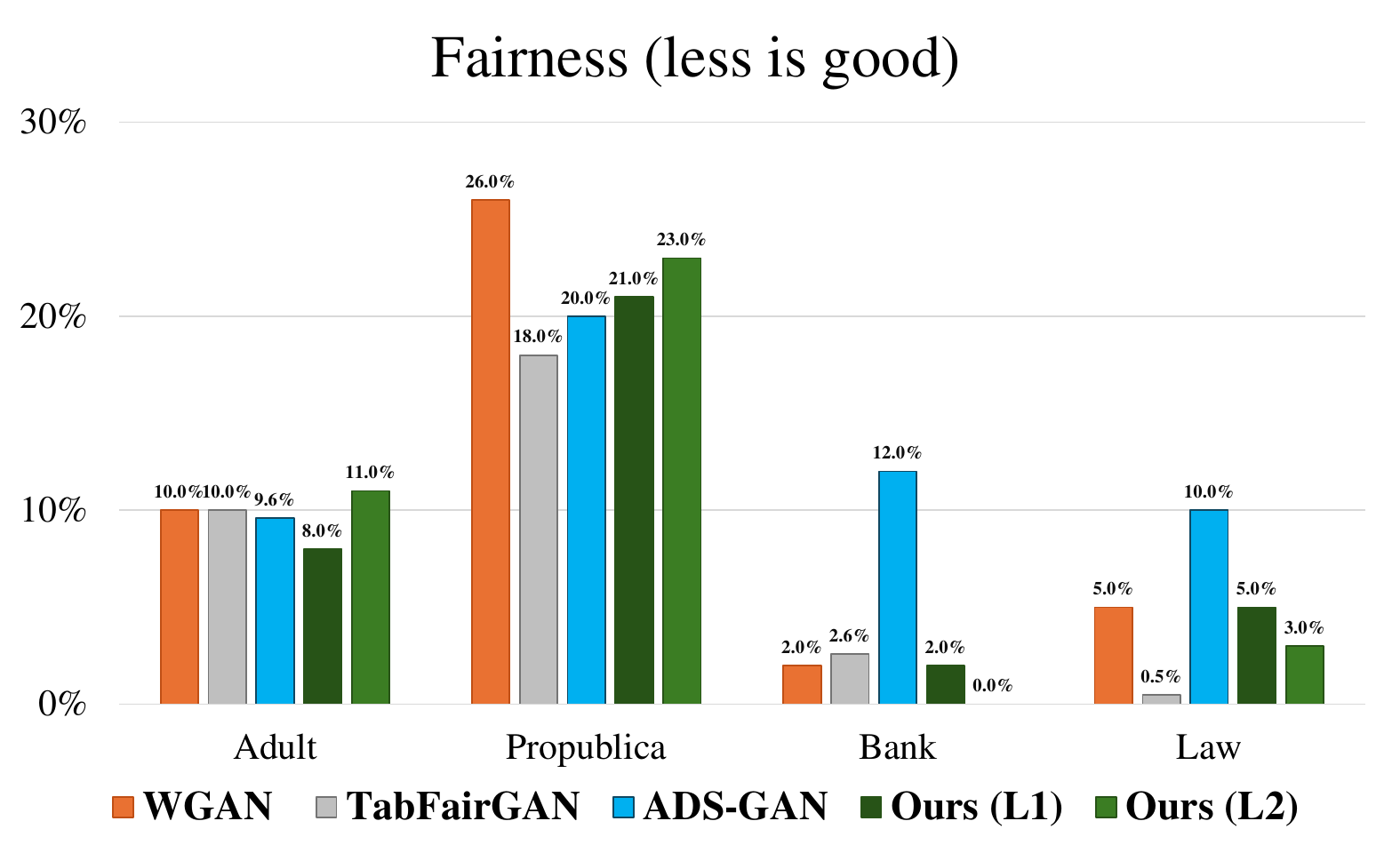}
  \caption{Result: Comparison among different models for fairness using checking demographic parity in generated synthetic data.}\label{fig:fair}
\end{figure}

We measure the ratio of favorable outcomes for sensitive and non-sensitive groups in the generated synthetic datasets by all models to calculate demographic parity for fairness. Figure \ref{fig:fair} shows the fairness performance by measuring demographic parity in the generated data across all models using four different datasets. Our model shows lower bias in the generated data for the Adult and Bank datasets among all models, while TabFairGAN performed well on the ProPublica and Law dataset. We also found that the privacy-preserving model ADS-GAN provides some level of fairness in the generated data, though less than other models. These experimental results indicate that our model effectively minimizes disparities between sensitive and non-sensitive groups compared to WGAN, TabFairGAN, and ADS-GAN in terms of fairness. 

\begin{figure}[t]
  \centering
  \graphicspath{{Figure/}}
  \includegraphics[width=1\linewidth]{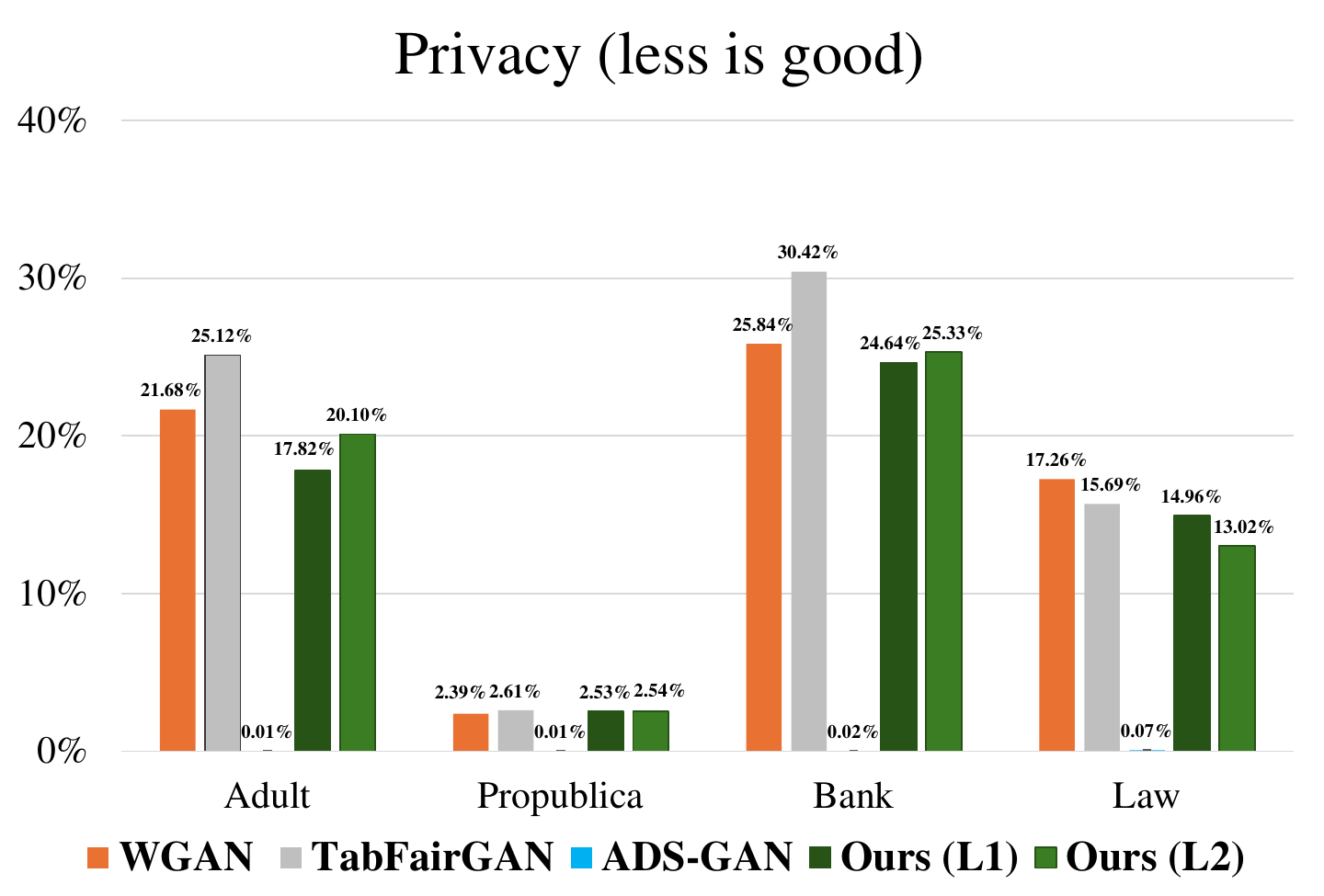}
  \caption{Result: Comparison among different models for privacy.}\label{fig:priv}
\end{figure}

To measure the re-identification risk, we calculated the identifiability score for all models. ADS-GAN excelled among all models, demonstrating its strong privacy-preserving capacity. Since we were also interested in evaluating how other models perform in terms of privacy, we measured the identifiability scores for data generated by WGAN and TabFairGAN, even though they are not privacy-focused models. Our model, PF-WGAN, outperformed both WGAN and TabFairGAN with significantly lower identifiability scores, suggesting a reduced risk of real data re-identification. For example, in the Adult dataset, the data generated by our model $(L1)$ had an identifiability score of 17.82\% ± 0.00\%, which is lower than the scores achieved by WGAN (21.68\% ± 0.00\%) and TabFairGAN (25.12\% ± 2.27\%). Figure \ref{fig:priv} shows the comparative privacy results between the models with all datasets. 

\textbf{Observation:} Our experiment reaffirms that without explicit privacy or fairness constraints, synthetic data generation models tend to produce less privacy-preserving and more biased data. Privacy-preserving model provide strong protection with loss of utility, though it offers limited fairness. On the other hand, fairness-focused model achieve good fairness but offer little privacy. However, when we incorporate both privacy and fairness into the generation process, it produces privacy-preserving, fair synthetic data. Overall, our model, PF-WGAN, effectively balances the dual objectives of fairness and privacy while maintaining competitive utility with lower standard deviations. This makes our model a robust and reliable solution for generating privacy-preserving, fair synthetic tabular data.
 
\section{Conclusion}\label{secConclusion}
Though privacy and fairness are two important concepts in synthetic data generation, no studies have evaluated these two concepts together in the literature. The aim of our research was to evaluate the performance of synthetic tabular data generation models in terms of privacy and fairness and to develop a solution for producing privacy-preserving, fair synthetic data. To achieve this, we developed a novel model, Privacy-Preserving Fair WGAN (PF-WGAN), by enhancing the WGAN model. The goal of this model was to generate synthetic tabular data that is both privacy-preserving and free from bias towards any particular group. For this purpose, we incorporated the identifiability score from ADS-GAN as a privacy loss function to ensure privacy in the generated data. We also employed a popular fairness measure, demographic parity, as a fairness loss metric. By integrating these privacy and fairness loss components into the traditional WGAN framework, we enhanced the model's ability to generate data that respects demographic parity and minimizes identifiability risks. Our approach does not require an additional generator or discriminator for data generation.

Through experimentation across multiple datasets, we found that without any privacy or fairness constraints, synthetic tabular data generation models offer limited privacy and fairness. In contrast, our model, PF-WGAN, offers more privacy and fairness while maintaining the usefulness of synthetic data. Our approach offers a promising solution for addressing bias in datasets while ensuring data privacy, paving the way for more ethical and responsible data-driven decision-making.

\textbf{Limitations and Future work:} In this study, we explore a new approach to generating privacy-preserving, fair synthetic tabular data by incorporating elements of privacy and fairness as the loss functions in the model's generator, which has shown promising performance. However, there is still room for improvement. While we opted for the \textit{identifiability} score from ADS-GAN to balance privacy and utility in this study, incorporating differential privacy into the PF-WGAN model, despite the potential utility loss due to added noise, could be a valuable future direction. Exploring how to effectively integrate differential privacy into the PF-WGAN framework while minimizing utility loss could lead to a more versatile model. Additionally, we did not explore attack-based evaluation methods such as membership inference, attribute inference, and linkage attacks. Evaluating the model's effectiveness in preserving privacy under different adversarial conditions using these methods could be pursued in the future. Furthermore, exploring other fairness metrics (e.g., equalized odds, disparate impact) could provide a more comprehensive assessment of the model's ability to mitigate bias. Finally, developing methods to incorporate multiple sensitive attributes or columns and allowing users to specify or combine their choice of fairness criteria could further enhance the model's flexibility and applicability in diverse scenarios.


{
\bibliographystyle{IEEEtran}
\bibliography{sample}
}


\end{document}